\documentclass[10pt,twocolumn,letterpaper]{article}

\usepackage{cvpr}              

\usepackage{multirow}

\definecolor{cvprblue}{rgb}{0.21,0.49,0.74}
\usepackage[pagebackref,breaklinks,colorlinks,allcolors=cvprblue]{hyperref}
\usepackage{cuted}
\usepackage[table]{xcolor}
\definecolor{neonyellow}{RGB}{255,255,153}
\definecolor{lightblue}{RGB}{204,229,255}

\def\paperID{4651} 
\def\confName{CVPR}
\def\confYear{2026}

\newcommand{\bench}  {RefBench-PRO\xspace}

\title{\bench: Perceptual and Reasoning Oriented Benchmark for Referring Expression Comprehension}

\author{
 \textbf{Tianyi Gao}$^{1,2}$\textsuperscript{*}\quad
 \textbf{Hao Li}$^{2,3}$\textsuperscript{*}\quad 
 \textbf{Han Fang}$^{2}$\quad
 \textbf{Xin Wei}$^{2}$\quad 
 \textbf{Xiaodong Dong}$^{2}$\quad \\
 \textbf{Hongbo Sun}$^{2}$ \quad  
 \textbf{Ye Yuan}$^{2}$\quad
 \textbf{Zhongjiang He}$^{2}$\quad
 \textbf{Jinglin Xu}$^{4}$\quad
 \textbf{Jingmin Xin}$^{1}$\textsuperscript{\dag}\quad
 \textbf{Hao Sun}$^{2}$\textsuperscript{\dag}\quad
\\ \\
$^1$National Key Laboratory of Human-Machine Hybrid Augmented Intelligence, \\
National Engineering Research Center for Visual Information and Applications, \\
Institute of Artificial Intelligence and Robotics, Xi’an Jiaotong University \\ \quad
$^2$Institute of Artificial Intelligence (TeleAI), China Telecom \\ \quad
$^3$Shanghai Jiao Tong University \quad
$^4$University of Science and Technology Beijing
}

\begin{document}
\maketitle
\begingroup
\renewcommand\thefootnote{*}%
\footnotetext{Equal Contributions.}%
\renewcommand\thefootnote{\dag}%
\footnotetext{Corresponding Authors.}%
\endgroup
\begin{abstract}
Referring Expression Comprehension (REC) is a vision-language task that localizes a specific image region based on a textual description. Existing REC benchmarks primarily evaluate perceptual capabilities and lack interpretable scoring mechanisms, which cannot reveal the grounding capability of Multi-modal Large Language Model (MLLM) across different cognitive abilities.
To address this limitation, we introduce \bench, a comprehensive  REC benchmark, which decomposes referring expressions into two core dimensions, i.e., perception and reasoning, and further subdivides them into six progressively challenging tasks, such as attribute, position, interaction, commonsense, relation and reject.
We also develop a fully automated data-generation pipeline that produces diverse referring expressions across these six sub-dimensions. 
Furthermore, We propose Ref-R1, an RL-based learning scheme, which incorporates Dynamic IoU-based GRPO to improve localization accuracy under increasingly complex reasoning conditions, establishing a stronger baseline for REC. 
Extensive experiments demonstrate that our \bench enables interpretable evaluation of MLLM on referring expression comprehension, presenting greater challenges in both perception and reasoning. Project page: \url{https://tg0110.github.io/prbench.github.io/}



\end{abstract}    

\begin{figure}
\centering 
\includegraphics[width=\columnwidth]{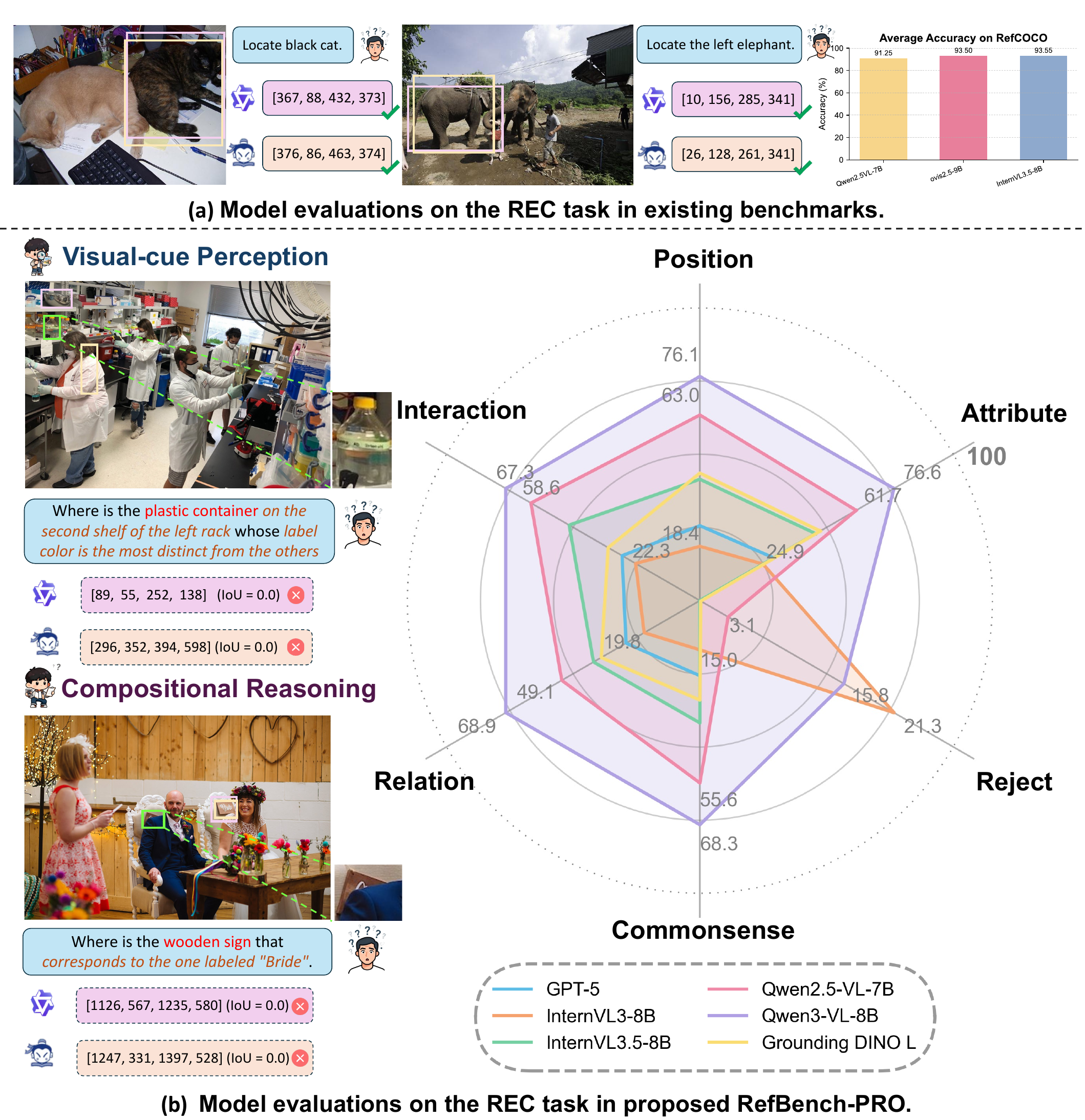}
\caption{Comparison of the proposed RefBench-PRO and the RefCOCO benchmark. (a) Existing REC benchmarks focus on images with dominant or few objects and use explicit, highly specific referring expressions, evaluated by a single aggregate metric. This design leads to near-saturation performance in MLLMs, obscuring their true capabilities in REC tasks. (b) We propose RefBench-PRO, a comprehensive and holistic benchmark across six critical dimensions.
By systematically revealing the specific limitations of current MLLMs in REC tasks, it provides new insights for designing MLLMs with improved localized understanding capabilities.}
\label{fig:motivation}
\end{figure}
\section{Introduction}

Referring Expression Comprehension (REC) ~\cite{liu2024grounding, bai2025univg, ma2024groma, jiang2024chatrex, jiang2025referring, jiang2025rex, shen2025vlm, li2025migician} is a fine-grained visual grounding task that requires models to process region-specific inputs  and localize the corresponding target region within an image. 
Due to limited vision-language understanding, specialized detection models such as Grounding DINO~\cite{liu2024grounding} struggle to accurately localize objects described by highly expressive and semantically complex referring expressions.
In contrast, multimodal large language models (MLLMs) ~\cite{gpt4v, bai2025qwen2, chen2024expanding, deitke2024molmo, wang2024qwen2,lu2024ovis,zhang2024mm1,li2024llava,li2025eagle}  have demonstrated remarkable generalization in both fine-grained perception and understanding, attracting growing interest in REC. 

While recent MLLMs have achieved promising performance on existing REC benchmarks, particularly on RefCOCO/+/g~\cite{yu2016modeling,yu2016modeling,mao2016generation}, 
They are still challenged by visually complex scenes or compositional vision-language reasoning. Several benchmarks are proposed to enable more robust evaluation. FineCops-Ref~\cite{liu2024finecops} introduces difficulty levels based on the complexity of  visual content, while Ref-L4~\cite{chen2025revisiting} emphasizes the construction of more detailed referring expressions.
However, these works still frame REC as a perceptual task, using more detailed descriptions but without incorporating compositional reasoning. Consequently, the lack of comprehensive task definitions and high-diversity visual contexts limits MLLMs’ ability to develop fine-grained and reasoning-aware visual grounding.

To address this gap, we propose \textbf{\bench}, a  holistic benchmark designed to comprehensively and interpretably  evaluate the perception and reasoning capabilities of MLLMs. First, the evaluation encompasses two core dimensions:  visual-cue perception and compositional reasoning.  Specifically, visual-cue perception is structured into three  aspects: (1) Attribute, which includes intrinsic visual properties; (2) Position, defined as spatial relationships  among different objects; and (3) Interaction,  referring to relative relationships among objects of the same category.  This dimension evaluates the model’s ability to accurately perceive relevant visual cues with the explicit  textual description. In contrast, compositional reasoning involves more challenging aspects: (4) Relation, which involves compositional referring that requires analyzing multiple objects; (5) Commonsense, which refers to objects via contextual descriptions rather than explicit naming; and (6) Rejection,  which handles referring to describing objects absent from the  image. Therefore, models must jointly reason over visual evidence  and textual cues to infer the correct referent.

Second, unlike conventional benchmarks that focus on dominant, few-object scenes, \bench employs high-resolution images with diverse object categories, targeting small, spatially scattered regions. This design provides fine-grained visual context while introducing a new challenge to accurately integrate various cues. Finally, \bench comprises six specific tasks, each containing 1,000 question-answer pairs, covering over 1,000 distinct objects,  with an average target object area ratio of 10\%.

Furthermore, based on the above design and to mitigate the risk of data leakage, we adopt FineHARD~\cite{xie2025fgclip} as our source data and propose a fine-grained referential annotation pipeline. This pipeline generates over 200k high-quality training pairs, named RefObjects-200k. To progressively improve the perception and reasoning ability of MLLMs, we employ Ref-R1, a two-stage training strategy. First, Chain-of-Thought based tuning is adopted to guide the model to capture text-related visual cues. Second, we propose Dynamic IoU-based Group Relative Policy Optimization (DyIoU-GRPO) to fuse visual and textual evidence into the reasoning chain, thereby enhancing perceptual localization and complex reasoning

Our contributions can be summarized as follows:
\begin{itemize}
    \item    We propose RefBench-PRO, a holistic benchmark comprising six critical aspects designed to evaluate the referring expression comprehension capabilities of MLLMs.
    \item We introduce a fine-grained referential annotation pipeline and present RefObjects-200k, a large-scale referring expression training set, alongside the training framework Ref-R1 to establish a competitive baseline. 
    \item We evaluate a range of widely used SOTA  MLLMs on RefBench-PRO to gain deeper insights into their limitations in fine-grained, reasoning-aware visual grounding.
\end{itemize}

\section{Related Work}



\begin{figure*}
\includegraphics[width=\linewidth]{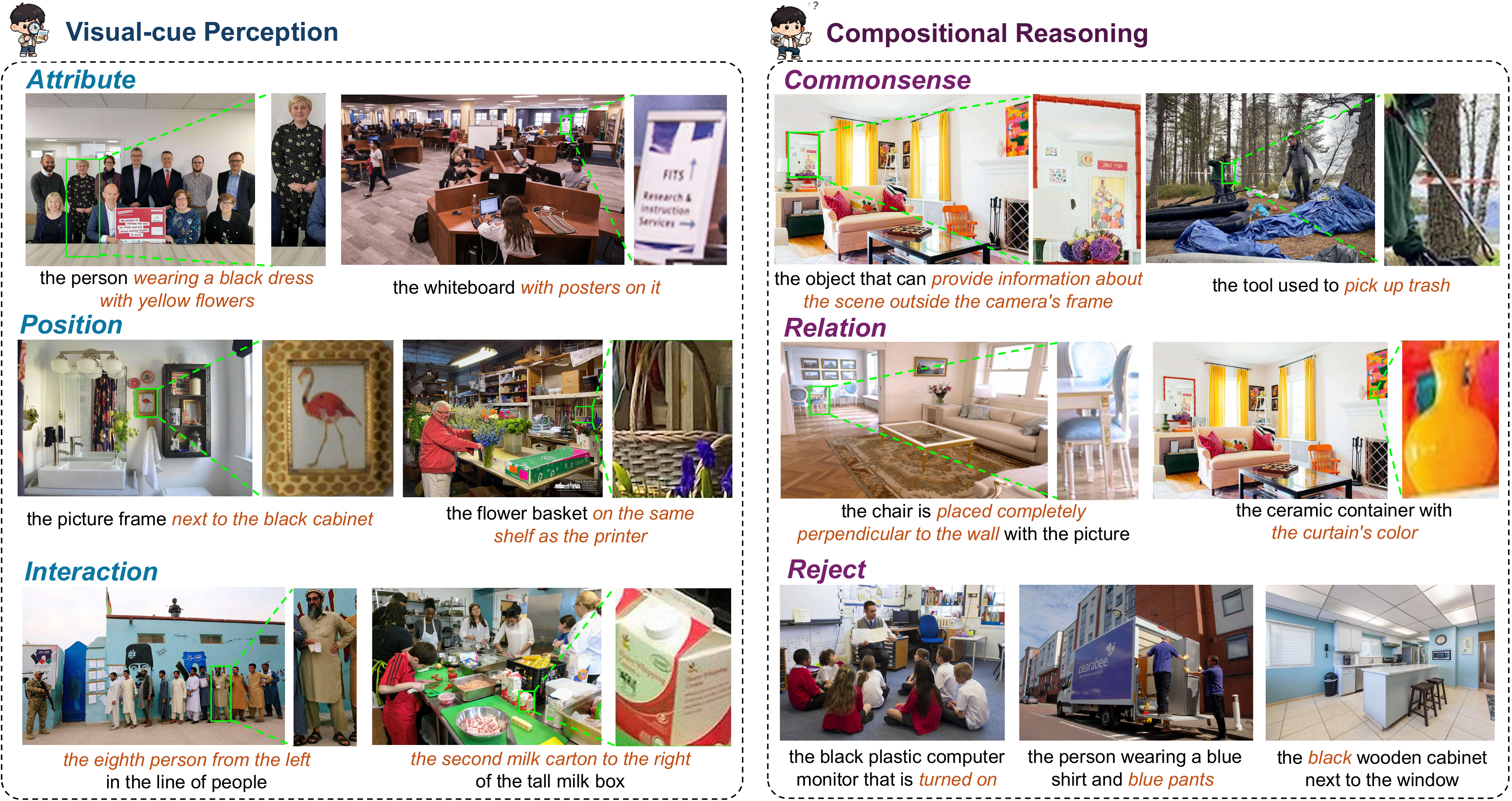}
\caption{Visualizations of the six tasks in the \bench. The three tasks on the left (Attribute, Position, Interaction) belong to Visual-cue Perception, while the three on the right (Commonsense, Relation, Reject) belong to Vision-language Interleaved Reasoning.} \label{fig:bench_exp}
\end{figure*}
 
\textbf{REC Benchmarks.} 
Referring Expression Comprehension (REC)~\cite{kazemzadeh2014referitgame,mao2016generation,zhang2019referring,luo2020multi,qiao2020referring,zheng2022towards} is the task of localizing a unique region in an image based on  the description. Widely adopted benchmarks such as RefCOCO~\cite{yu2016modeling}, RefCOCO+~\cite{yu2016modeling}, and RefCOCOg~\cite{mao2016generation} are built upon MSCOCO~\cite{lin2014microsoft}, leveraging its dense annotations.  These benchmarks extend MSCOCO~\cite{lin2014microsoft} with diverse referring expressions,  establishing REC as a standard benchmark for evaluating the perception capability of MLLMs. However, performance on these benchmarks has saturated due to limited visual diversity, driving recent work toward harder, more diverse challenges. FineCops-Ref~\cite{liu2024finecops} introduced difficulty levels based on the complexity of content. Ref-L4~\cite{chen2025revisiting} introduced more descriptive expressions. C-REC~\cite{yu2024revisiting} proposed counterfactual negative examples to increase visual ambiguity. GREC~\cite{he2023grec} tackled the multi-target REC task. Migician~\cite{li2025migician} introduced a multi-image grounding task requiring multi-context understanding.
However, these benchmarks still treat REC as a multimodal perception task and lack interpretable metrics to diagnose models' fine-grained understanding of local regions. 
Instead, we propose \bench to evaluate models’ perception and reasoning capabilities jointly.

\noindent\textbf{MLLM-based REC Methods.}
Recent advances in MLLMs~\cite{gpt4v, bai2025qwen2, wu2024deepseek, chen2024expanding, alayrac2022flamingo, deitke2024molmo, wang2024qwen2, lu2024ovis, zhang2024mm1, li2024llava, li2025eagle} have demonstrated promising grounding capabilities through the fusion of visual and textual content. 
To further strengthen these capabilities, recent works have explored several methods.
Groma~\cite{ma2024groma}, ChatRex~\cite{jiang2024chatrex}, and RexSeek~\cite{jiang2025referring} reformulated REC as a region-token retrieval task using an additional specialist model.
PAM~\cite{lin2025perceive} extends the SAM 2 framework with LLM to support both semantic understanding and visual perceptual capabilities.
Visual CoT~\cite{shao2024visual} utilizes CoT to fully leverage the reasoning capabilities of MLLMs.
Reinforcement learning, such as GRPO~\cite{shao2024deepseekmath}, has been employed to further enhance complex reasoning capabilities~\cite{yu2025perception, liu2025visual, bai2025univg, shen2025vlm}.
VLM-R1~\cite{shen2025vlm} and Visual-Rft~\cite{liu2025visual} enhance model capabilities by adjusting the reward function. Deepeyes~\cite{shen2025vlm} and Chain-of-Focus~\cite{liu2025visual} introduce an image zoom-in tool to capture visual details.
However, existing methods have not sufficiently explored compositional referring expressions. To this end, we propose a fine-grained annotation pipeline with RL-based framework, Ref-R1, to establish a more comprehensive baseline for REC.

\section{\bench}
\bench is a holistic benchmark for Referring Expression Comprehension  (REC), designed to offer diverse evaluation dimensions and challenging  scenarios. As shown in Figure~\ref{fig:bench_exp}, it focuses on evaluating two critical capabilities of  MLLMs: \textbf{Visual-cue Perception} and \textbf{Compositional Reasoning},  each further subdivided into three subcategories. Visual-cue perception requires models to capture sufficient visual evidence given specific referring queries. Beyond that, Compositional Reasoning demands to integrate implicit textual references with contextual visual relations.  

\subsection{\bench Construction}
\begin{figure*}[htbp]
\includegraphics[width=\linewidth]{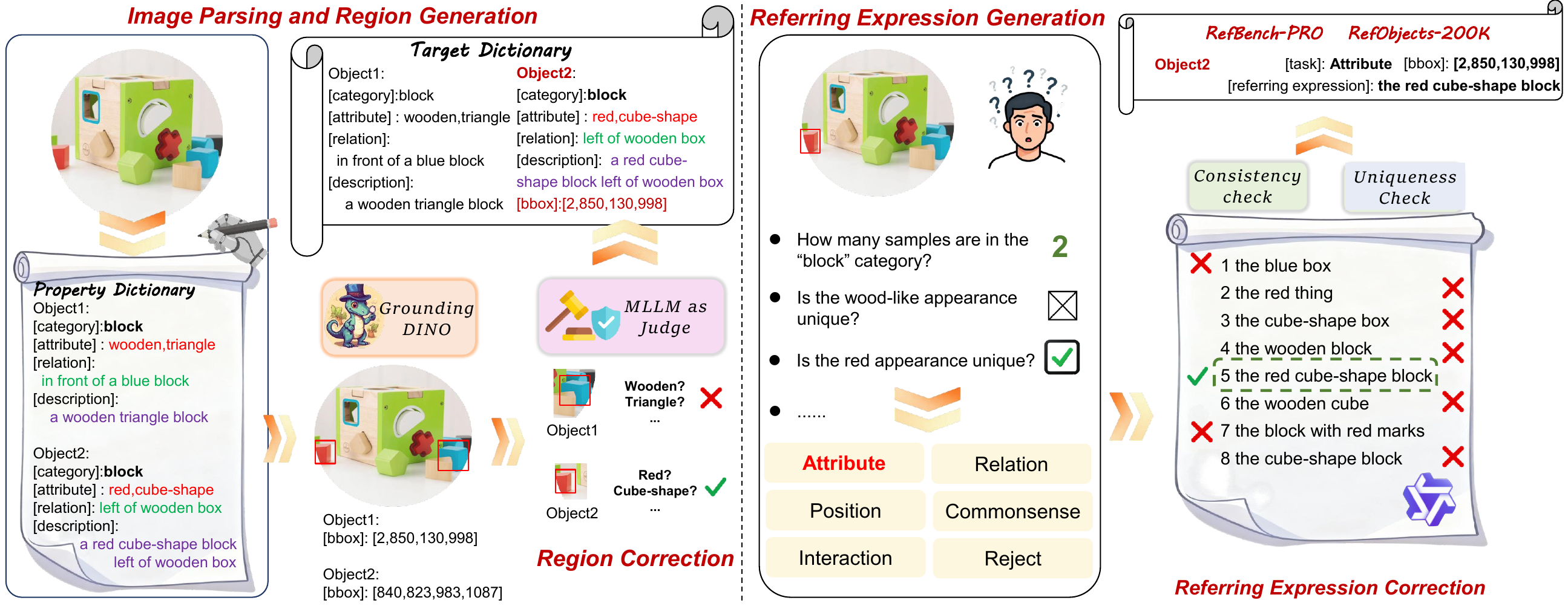}
\caption{The fine-grained referential annotation pipeline to construct RefObjects-200k and \bench.} 
\label{fig:data_construct_pipeline}
\end{figure*}

\label{sec:data_generation_pipeline}

To construct \bench, we leverage FineHARD~\cite{xie2025fgclip}, a  large-scale dataset annotated with bounding boxes across a wide range of  resolutions. Source images are filtered based on resolution variability  and the distribution of Grounding DINO-detected bounding boxes,  retaining only those exhibiting high resolution, multiple diverse  objects, and rich visual content. To generate high-quality query–bounding box  pairs across six tasks, we propose the fine-grained annotation pipeline as depicted in Figure~\ref{fig:data_construct_pipeline}:

\noindent\textbf{Image Parsing and Region Generation.} For each image $I$, we prompt Qwen2.5-VL-72B to generate a structured property dictionary encoding the visual elements, including the object category $c$, inherent attributes  $a$, interactive relationships with other objects $r$, and a brief description phrase $d$. Then, we use Grounding DINO~\cite{liu2024grounding} to ground the description phrase $d$ to obtain the bounding boxes $b$. To ensure visual complexity, we retain images containing at least two distinct object categories and five  total objects. Consequently, the property for each image $I$  is structured as:
\begin{equation}
    \mathcal{E_{I}} = \{(a_j,b_j,c_j,d_j,r_j)\}_{j = 1}^{N},\, \text{with} \, |c_j| > 2, N \geq 5,
\label{Parsing}
\end{equation}
where $j$ indicates the index of specific object.

\noindent\textbf{Region Correction.} For each object $j$, we convert each element of $\mathcal{E_{I}}$  into a verification checklist. Qwen2.5-VL-72B is then prompted to evaluate whether the visual content within the target region aligns with the corresponding property in the checklist. Only objects that are fully consistent across all properties are retained, as they provide reliable  spatial coordinates and diverse property descriptions.





 



\noindent\textbf{Referring Expression Generation.} Once the properties are verified, we design a task selection mechanism to automatically assign each retained object to a specific task type, formalized by the selection function $\mathcal{F}_{\text{select}}$  as follows: 
\begin{equation}
T_j = \mathcal{F}_{\text{select}}(t_j, \mathcal{E}_{I}).
\end{equation}
where $t_j$  is the target object to be assigned. $\mathcal{F}_{\text{select}}$ is a human-defined rule-based selection function designed to identify challenging instances for each task by defining  specific visual conditions. For example, we assign objects that belong to the same category but exhibit distinct inherent attributes to \textit{Attribute} task, providing more fine-grained visual cues as the priors. These rules intentionally introduce visual complexity, thereby curating a set of high-quality, task-relevant seed objects. Then we prompt Qwen2.5-VL-72B to reformulate the property to generate candidate expressions for the assigned object in specific task $T_j$.

\noindent\textbf{Referring Expression Correction.} Each referring expression is validated through a two-stage verification process. First, a consistency check is adopted to verify that the visual content within the bounding box semantically aligns with the expression, filtering out hallucinated descriptions.
Second, a uniqueness check ensures that the expression unambiguously refers to a single, distinct region in the image. \textbf{More details regarding the  dataset construction will be provided in the supplemental material.}

\begin{table}[]
\centering
\resizebox{1\columnwidth}{!}{%
\begin{tabular}{lccccccc}
\toprule
\multicolumn{8}{c}{\textbf{\bench}}                                                                                                                                                       \\ \midrule
\multicolumn{1}{l|}{type}             & Attribute & Position & Interaction & Relation & Commonsense & \multicolumn{1}{c|}{Reject} & Total   \\ \midrule
\multicolumn{1}{l|}{Images}           & 701       & 735      & 642         & 695      & 715         & \multicolumn{1}{c|}{764}    & 3102   \\
\multicolumn{1}{l|}{Phrases}       & 1,000     & 1,000   & 1,000  & 1,000   & 1,000 & \multicolumn{1}{c|}{1,000}  & 6,000   \\
\multicolumn{1}{l|}{ Img Size}    & 1602x1301 & 1588x1294& 1608x1292   & 1619x1282& 1627x1280   & \multicolumn{1}{c|}{1629x1301}& 1612x1292 \\
\multicolumn{1}{l|}{ Box Size}  & 10.76\%   & 8.34\%   & 9.43\%     & 6.15\%   & 6.14\%      & \multicolumn{1}{c|}{N/A}    & 8.16\%  \\
\multicolumn{1}{l|}{Length}  & 9.3       & 11.0     & 10.3        & 14.4     & 15.5        & \multicolumn{1}{c|}{13.8}     & 12.4    \\ \bottomrule

\multicolumn{8}{c}{\textbf{RefObjects-200k}}                                                                                                                                                           \\ \midrule
\multicolumn{1}{l|}{type}             & Attribute & Position & Interaction & Relation & Commonsense & \multicolumn{1}{c|}{Reject} & Total   \\ \midrule
\multicolumn{1}{l|}{Images}           & 14,284     & 11,083    & 7,696       & 7,945    & 21,055      & \multicolumn{1}{c|}{18,670}  & 28,541  \\
\multicolumn{1}{l|}{Phrases}       & 37,257    & 24,380   & 16,903      & 17,193   & 59,106      & \multicolumn{1}{c|}{49,146} & 203,985  \\
\multicolumn{1}{l|}{Img Size}    & 1629x1274 & 1636x1275& 1603x1287   & 1633x1268& 1605x1295   & \multicolumn{1}{c|}{1608x1294}& 1616x1286 \\
\multicolumn{1}{l|}{ Box Size}  & 11.27\%   & 8.66\%  & 9.54\%     & 8.33\%   & 8.27\%      & \multicolumn{1}{c|}{N/A}    & 9.68\%  \\
\multicolumn{1}{l|}{Length}  & 8.6       & 9.3      & 9.3         & 15.3     & 16.2        & \multicolumn{1}{c|}{14.3}   & 12.88    \\ \midrule

\end{tabular}%
}
\caption{The distribution of RefBench-PRO and RefObjects-200k.}
\label{tab:dataset_statistics}
\end{table}

\begin{figure}[!t]
    \centering
    \begin{subfigure}[b]{0.49\columnwidth}
        \centering
        \includegraphics[width=\columnwidth]{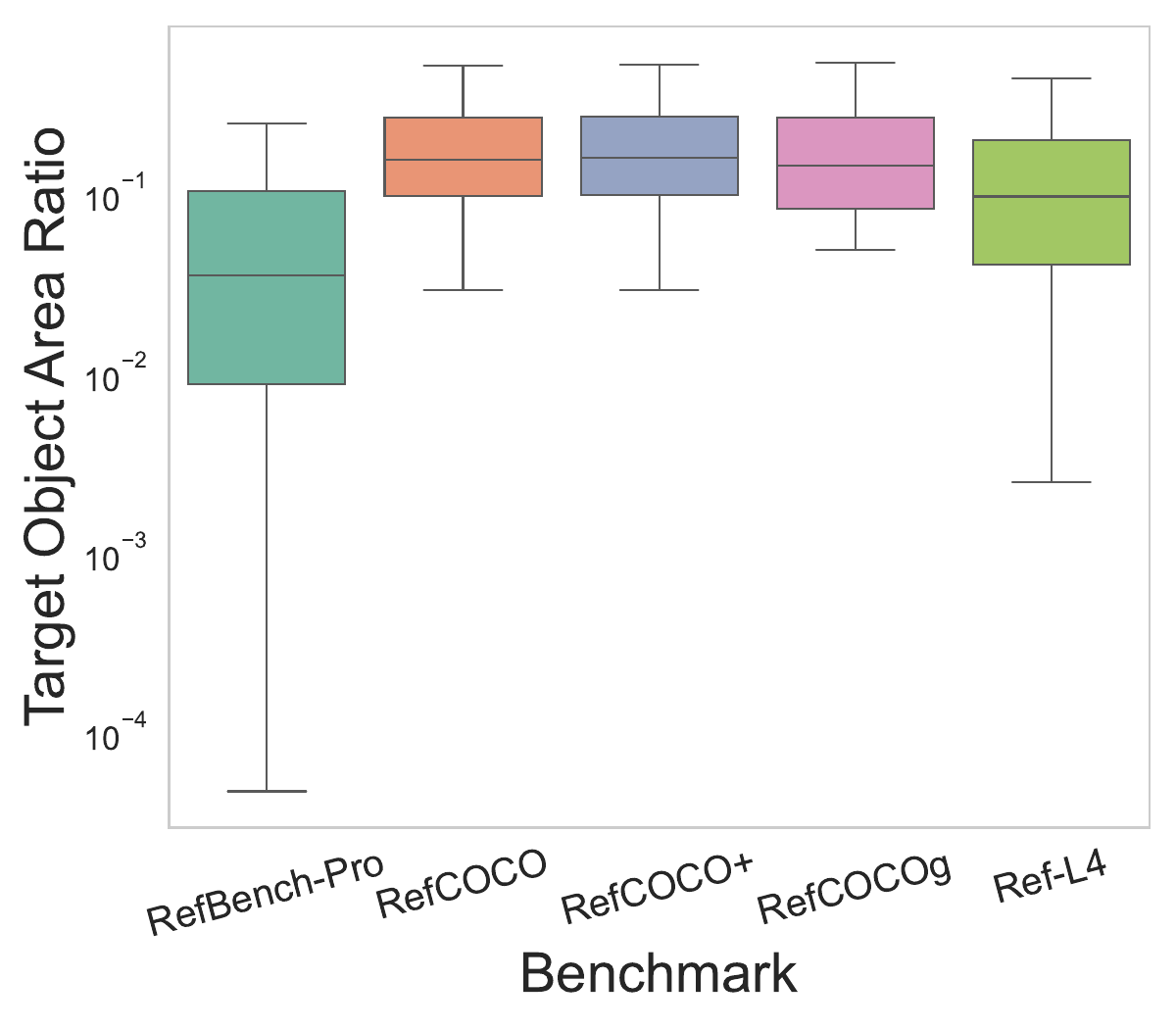}
        \caption{}
        \label{fig:bbox_compare}
    \end{subfigure}
    \begin{subfigure}[b]{0.49\columnwidth}
        \centering
        \includegraphics[width=\columnwidth]{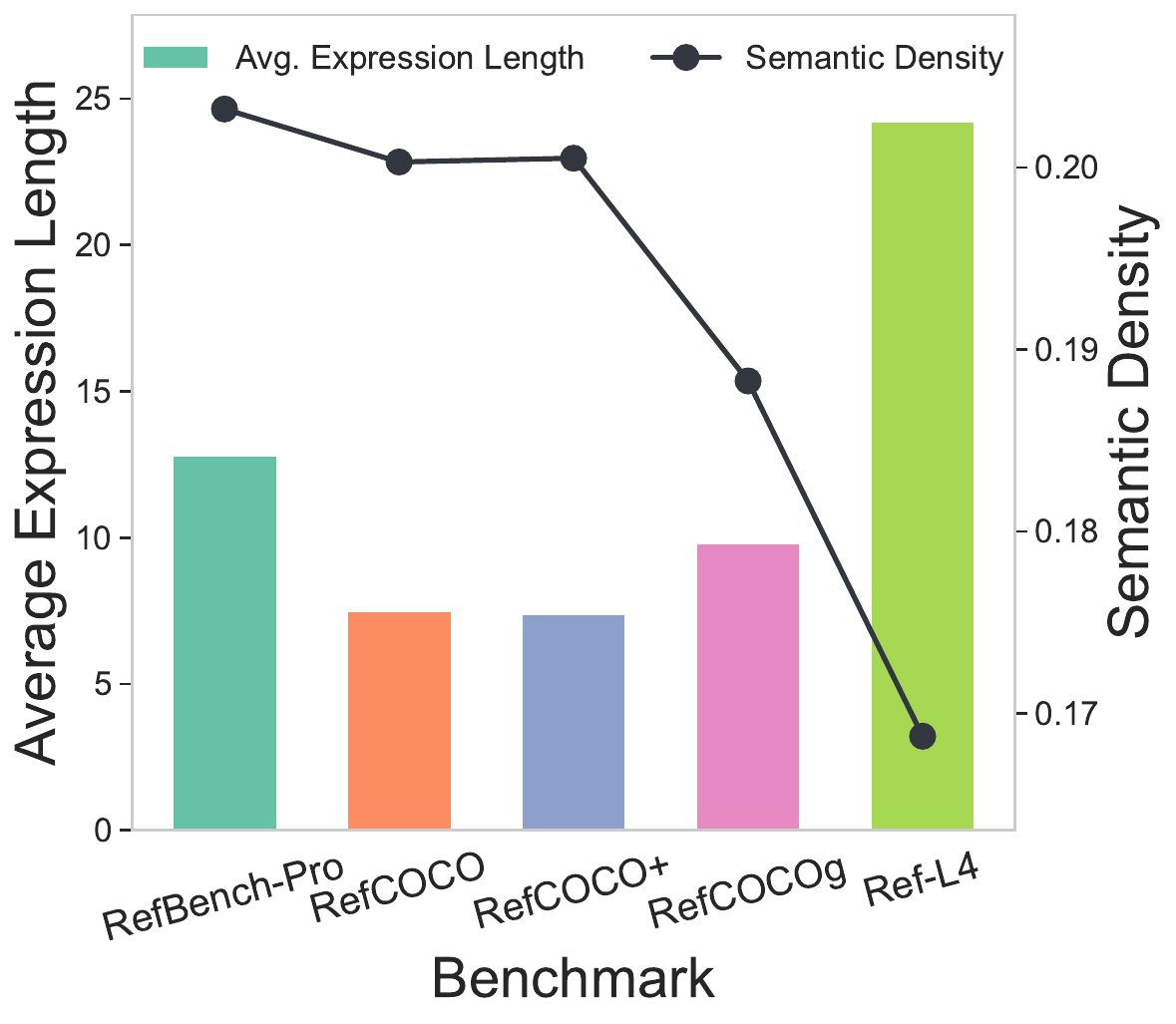}
        \caption{}
        \label{fig:semantic_complexity}
    \end{subfigure}
    \caption{A comparative analysis of \bench against established benchmarks (RefCOCO~\cite{yu2016modeling}, RefCOCO+~\cite{yu2016modeling}, RefCOCOg~\cite{mao2016generation}, and Ref-L4~\cite{chen2025revisiting}) evaluates (a) target object area ratios and (b) average expression length and semantic density, defined as the number of visual cues per word.}
    \label{fig:analysis}
\end{figure}

\subsection{Data statistics}
We construct RefObjects-200k via our annotation pipeline. A subset of pairs across six dimensions is then sampled, further corrrected by ten annotators to establish \bench. Additional details are provided in Table \ref{tab:dataset_statistics}. 
Specifically, REFbench-PRO contains 6,000 pairs distributed across six sub-categories: Attribute, Position, Interaction, Relation, Commonsense, and Rejection. The benchmark features high-resolution images, covers over 1,000 distinct object types, and emphasizes small or marginally visible targets, with an average target object area ratio of 10\%. In Figure~\ref{fig:bbox_compare}, our RefBench-PRO exhibits a broader distribution, with greater emphasis on objects with small relative size. Additionally, we depict object density and find that Ref-L4~\cite{chen2025revisiting} generates longer expressions containing context-irrelevant content, whereas our \bench produces comparably long descriptions by incorporating discriminative visual cues, achieving higher information density.

\section{Methodology}
\begin{table*}[t]
\centering
\caption{Performance of state-of-the-art models on the \bench across six aspects. The best results are in bold, and the second-best results are underlined.}
\resizebox{1.0\linewidth}{!}{
\begin{tabular}{l c cc cccc cccc}
\toprule
\multirow{2}{*}{\textbf{Models}}
& \multirow{2}{*}{\textbf{Size}}
& \multicolumn{2}{c}{\textbf{Overall}}
& \multicolumn{4}{c}{\textbf{Visual-cue Perception}} & \multicolumn{4}{c}{\textbf{Compositional Reasoning}} \\
\cmidrule(lr){3-4} \cmidrule(lr){5-8} \cmidrule(lr){9-12}
 & &\textbf{$\text{Acc}_p$} & \textbf{$\text{Acc}_o$} & \textbf{Attribute} & \textbf{Position} & \textbf{Interaction} & \textbf{$\text{Acc}_{API}$} & \textbf{Relation} & \textbf{Commonsense} & \textbf{Reject} & \textbf{$\text{Acc}_{RC}$} \\
\midrule
\rowcolor{red!10}\multicolumn{12}{c}{\textbf{Open-vocabulary Grounding Models}} \\
GLEE~\cite{wu2024general}& - &36.1 & 31.2 & 48.2 & 38.4 & 34.5 & 40.4 & 31.4&27.9&7.1 & 29.7 \\
Grounding DINO L~\cite{liu2024grounding}& - &37.6 & 31.3 & 47.5 & 43.3 & 31.8 & 40.9 & 35.0 & 30.3 & 0.1 & 32.7 \\

\rowcolor{green!10}\multicolumn{12}{c}{\textbf{Proprietary MLLMs}} \\
Gemini-2.5-pro~\cite{comanici2025gemini}& - & 9.6 & 8.0 & 10.4 & 11.5 & 10.8 & 10.9 & 7.2 & 8.2 & - & 7.7\\
GPT-4o~\cite{hurst2024gpt}& - & 12.1 & 10.1 & 11.7 & 12.8 & 11.9 & 12.1 & 12.4 & 11.6 & - & 12.0 \\
GPT-5~\cite{openai2025gpt5}& - & 26.1 & 21.8 & 29.2 & 25.5 & 27.0 & 27.2 & 26.2 & 22.9  & - & 24.6 \\

\rowcolor{blue!10}\multicolumn{12}{c}{\textbf{Specialist MLLMs}} \\
PaDT~\cite{su2025patch}&  3B   & 26.6 & 22.2 & 30.4 & 28.0 & 30.4 & 29.6 & 23.7 & 20.8 & - & 22.3\\
VLM-R1~\cite{shen2025vlm}& 3B & 54.4 & 45.3 & 59.0 & 58.0 & 54.1 & 57.0 & 47.8 & 53.2 & - & 50.5 \\
ChatRex~\cite{jiang2024chatrex}&  7B  & 49.5 & 41.3 & 54.7 & 51.1 & 53.2 & 53.0 & 45.1 & 43.4 & -  & 44.2 \\
Migician~\cite{li2025migician}& 7B & 52.3 & 43.6 & 57.3 & 59.7 & 52.8 & 56.6 & 45.4 & 46.1 & - & 45.7 \\
UniVG-R1~\cite{bai2025univg}& 7B & 53.0 & 44.2 & 59.4 & 57.2 & 55.1 & 57.2 & 48.3 & 44.9 & - & 46.6 \\
Rex-Thinker~\cite{jiang2025rex} & 7B & 63.6 & 53.0 & 67.1 & 64.5 & 61.7 & 64.4 & 59.3 & \underline{65.6} & - & 62.4\\
CogVLM-Grounding~\cite{wang2024cogvlm}& 17B & 57.1 & 47.5 & 62.4 & 62.4 & 55.9 & 60.2 & 49.4 & 55.2 & - & 52.3 \\

\rowcolor{yellow!15}\multicolumn{12}{c}{\textbf{Open-source General MLLMs}} \\
Qwen2-VL~\cite{wang2024qwen2}& 7B & 45.4 & 42.6 & 55.3 & 47.8 & 37.8 & 47.0 & 39.4 & 46.5 & \textbf{28.5} & 43.0\\
Mimo-VL-RL~\cite{xiaomi2025mimo}& 7B & 56.3 & 46.9 & 60.9 & 58.4 & 57.3 & 58.9 & 51.8 & 52.9 & 0.1 & 52.4\\
Qwen2.5-VL~\cite{bai2025qwen2}& 7B & 57.6 & 48.5 & 61.7 & 63.0 & 58.6 & 61.1 & 49.1 & 55.6 & 3.1 & 52.3\\
InternVL3~\cite{zhu2025internvl3}& 8B & 20.1 & 20.3 & 24.9 & 18.4 & 22.3 & 21.9 & 19.8 & 15.0 & 21.3 & 17.4\\
InternVL3.5~\cite{wang2025internvl3}& 8B & 41.5 & 34.6 & 45.7 & 41.2 & 45.3 & 44.1 & 37.8 & 37.3 & - & 37.5\\
LLaVA-OneVision-1.5~\cite{an2025llava}& 8B & 50.7 & 42.3 & 54.5 & 54.0 & 48.4 & 52.3 & 48.1 & 48.6 & - & 48.3\\
Qwen3-VL~\cite{bai2025qwen3}& 8B & \textbf{71.4} & \textbf{62.2} & \textbf{76.6} & \textbf{76.1} & \underline{67.3} & \textbf{73.3} & \textbf{68.9} & \textbf{68.3} & 15.8 & \textbf{68.6}\\
Ovis2.5~\cite{lu2025ovis2}& 9B & 61.7 & 51.5 & 65.7 & 63.6 & 59.7 & 63.0 & 58.7 & 61.0 & - & 59.9\\
GLM-4.1V-Base~\cite{vteam2025glm45vglm41vthinkingversatilemultimodal} & 9B & 60.1 & 50.1 & 62.9 & 61.0 & 57.7 & 60.5 & 57.0 & 61.9 & - & 59.4 \\
LLaVA-OneVision~\cite{li2024llava}&72B&56.5&47.1&60.1&59.4&53.7&57.7&54.2&55.0&-&54.6\\
Qwen2.5-VL~\cite{bai2025qwen2}& 72B & \underline{66.7} & \underline{59.5} & \underline{68.6} & \underline{69.1} & \textbf{69.4} & \underline{69.1} & \underline{61.6} & 64.8 & \underline{23.6} & \underline{63.2}\\
InternVL3~\cite{zhu2025internvl3} & 78B & 21.8 & 22.3 & 35.0 & 24.9 & 28.2 & 29.4 & 24.3 & 20.8 & 24.8 & 22.5\\
\bottomrule
\end{tabular}}
\label{tab:benchmarking}
\end{table*}


RefBench-PRO, a comprehensive REC benchmark that emphasizes visually cluttered scenes and compositional reasoning, reveals new challenges in the visual grounding capabilities of MLLMs. To address these challenges, we propose \textbf{Ref-R1}, a training framework designed to establish a foundational baseline for future research.

\subsection{Chain-of-Thought Cold Start.} 
Using the RefObjects-200k dataset, we leverage Qwen2.5-VL-72B~\cite{bai2025qwen2} to generate chain-of-thought reasoning for each sample. Guided by a predefined prompt, the model produces multiple reasoning paths, from which the highest-quality chain is selected as the final output. We curate 180K high-quality reasoning traces for first-stage supervised fine-tuning, yielding a cold-start model capable of generating bounding boxes through coherent, step-by-step reasoning.

\subsection{DyIoU-GRPO}
In the second stage, to improve generalization by exploring diverse reasoning trajectories, we propose Dynamic IoU-based GRPO (DyIoU-GRPO) for post-training. For each query $q$, GRPO samples a set of $N$  reasoning responses. For each response $o_i$, the rule-based reward $R_i = R(q,o_i)$ is computed to derive group-relative advantages:
\begin{equation}
    A_i = \frac{R_i - \text{mean}(\{R_1, R_2, \dots, R_N\})}{\text{std}(\{R_1, R_2, \dots, R_N\})}.
\end{equation}
The policy $\pi_\theta$ is then optimized to maximize the expected advantage as follows:
\begin{equation}
\begin{split}
        \mathcal{J}&_{GRPO}(\theta) = \mathbb{E}_{q \sim P(Q), \{o_i\}_{i=1}^N \sim \pi_{\theta_{old}}(O|q)} \\
    &\left[\frac{1}{N}\sum_{i=1}^N \frac{\pi_{\theta}(o_i|q)}{\pi_{\theta_{old}}(o_i|q)}A_i - \beta\mathbb{D}_{KL}(\pi_{\theta}||\pi_{ref})\right],
\end{split}
\end{equation}
where $\beta$ is a hyperparameter that controls the KL divergence penalty. 
Inspired by human cognition, where observers first perform a global scan to identify coarse regions of interest and then shift attention to explore fine-grained details, we design a hierarchical attention-based reward function to guide models in step-by-step reasoning, progressively integrating perception and reasoning as:

\noindent \textbf{Format Reward ($R_{format}$).} This reward ensures that the response strictly conforms to the required format as `$<$think$>$reasoning process$<$/think$>$$<$answer$>$[x1, y1, x2, y2]$<$/answer$>$'. It assigns a value of 1 if the format is correct and 0 otherwise.

\noindent \textbf{Dynamic IoU Reward ($R_{DyIoU}$).} 
Given the ground-truth bounding box coordinates $B_{GT}$ and the model’s predicted coordinates $B_{pred}$, we compute the IoU score as $IoU(B_{GT}, B_{pred})$.
The prediction is assigned a reward of 1 if its IoU score exceeds the threshold, and 0 otherwise. The threshold is defined as follows: 
\begin{equation}
    \resizebox{0.9\columnwidth}{!}{
    $\tau_{IoU}(t, s) = \max \left(  \alpha + (\beta - \alpha) \cdot \frac{t}{T} - d_{\text{max}} \cdot (1 - s), \, \alpha \right)$,
    }
\end{equation}
where $\alpha$ and $\beta$ represent the start and end threshold values. $t$ indicates the current training step, ranging from $0$ to $T$. $s$ denotes the ratio of the bounding box area to the total image area. $d_{\text{max}}$ controls the threshold penalty. 
As training progresses, the threshold increases to raise localization difficulty and encourage precise spatial predictions. Meanwhile, smaller targets receive lower thresholds to promote attention to fine-grained visual cues.

\noindent \textbf{Group Quality Reward.} To enhance discriminative group-relative advantages, we introduce the Group Quality Reward. We define a dynamic quality threshold $\tau_q$ that gradually increases during training to identify hard groups, where the number of correct responses is below $\tau_q$. For each hard group, the Group Quality Reward is computed as $\frac{k}{n}$, where $n$ is the group size and $k$ is the number of correct responses. This value is added to the reward of every correct response and subtracted from that of every incorrect response. By widening the reward gap within the group, this adjustment strengthens the model’s preference for correct answers, especially in challenging cases with high error rates.

Overall, the final reward $R$ is formulated as: 

\begin{equation}
R_i = 
\begin{cases} 
    R_{\text{format}} + R_{\text{DyIoU}} \pm \frac{k}{n} \cdot p & \text{if } n_{\text{correct}} < \tau_q,
    \\[2.5ex]
    R_{\text{format}} + R_{\text{DyIoU}} & \text{if } n_{\text{correct}} \ge \tau_q,
\end{cases}
\end{equation}
where $p$ is the weight of $\frac{k}{n}$ and the sign of the $\pm$ term is positive for a correct response $o_i$ and negative otherwise.

\section{Experiments}
\subsection{Experimental Settings}
 \textbf{Models and Baselines.} To comprehensively evaluate our proposed benchmark, we conduct experiments across 24 representative models, categorized into four groups: Open-vocabulary Grounding Models, Proprietary MLLMs, Specialist MLLMs, and Open-source General MLLMs.
The open-vocabulary grounding models include Grounding Dino L~\cite{liu2024grounding} and GLEE~\cite{wu2024general}.
The proprietary MLLMs comprise GPT-4o~\cite{hurst2024gpt}, GPT-5~\cite{openai2025gpt5}, and Gemini-2.5-Pro~\cite{comanici2025gemini}.
We further evaluate a set of specialist MLLMs tailored for REC tasks, including PaDT~\cite{su2025patch}, Vlm-R1~\cite{shen2025vlm}, ChatRex~\cite{jiang2024chatrex}, Migician~\cite{li2025migician}, UniVG-R1~\cite{bai2025univg}, Rex-Thinker~\cite{jiang2025rex} and CogVLM~\cite{wang2024cogvlm}.
Finally, we incorporate several general-purpose MLLMs, such as Qwen series~\cite{wang2024qwen2, bai2025qwen2, bai2025qwen3}, InternVL series~\cite{zhu2025internvl3, wang2025internvl3}, Mimo-VL~\cite{xiaomi2025mimo}, LLaVA-OneVision-1.5~\cite{an2025llava} and Ovis2.5~\cite{lu2025ovis2}. For all baseline methods, we conducted experiments based on their official code.

 \textbf{Evaluation Metrics.} We evaluate performance using two metrics: rejection accuracy (RejAcc) and mean accuracy (mAcc). For the \emph{Reject} category, a prediction is correct if the model outputs no bounding box or explicitly states that the target is absent. For the other five tasks, mAcc is computed as the average accuracy across IoU thresholds from 0.5 to 0.9 in steps of 0.05. To provide a overall assessment, we report additional metrics: $\mathrm{Acc}_o$, the mean accuracy over all six tasks; $\mathrm{Acc}_p$, the mean accuracy over the five non-rejection tasks; $\mathrm{Acc}_{API}$, the mean accuracy over the three perception tasks; and $\mathrm{Acc}_{RC}$, the mean accuracy over Relation and Commonsense.


 \textbf{Training Details.} We conduct experiments on Qwen-2.5-VL-7B~\cite{bai2025qwen2}. In the first stage, we train on 180K samples generated from RefObjects-200k for one epoch, using a learning rate of $5 \times 10^{-6}$ and an accumulated batch size of 16. In the second stage, we select 80k samples and train for 5,000 steps with a learning rate of $1 \times 10^{-6}$ and an accumulated batch size of 16. For reward design, we set $\alpha = 0.5$, $\beta = 0.8$, and $d_{\text{max}} = 0.15$ for the dynamic IoU reward, and $p = 0.5$ for the group quality reward.  All experiments are conducted on 8 NVIDIA A800-80G GPUs.

 \begin{figure*}[t]
    \centering
    \includegraphics[width=\linewidth]{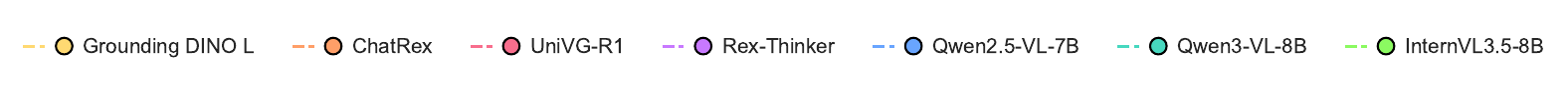} \\
    \begin{subfigure}{0.33\textwidth}
        \centering
        \includegraphics[width=\linewidth]{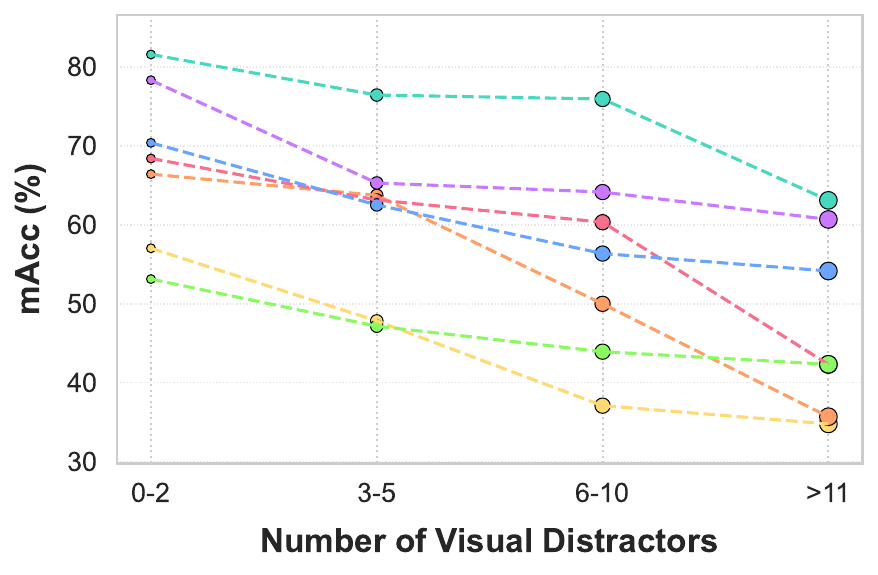}
        \caption{}
        \label{fig:same_category}
    \end{subfigure}
    \begin{subfigure}{0.33\textwidth}
        \centering
        \includegraphics[width=\linewidth]{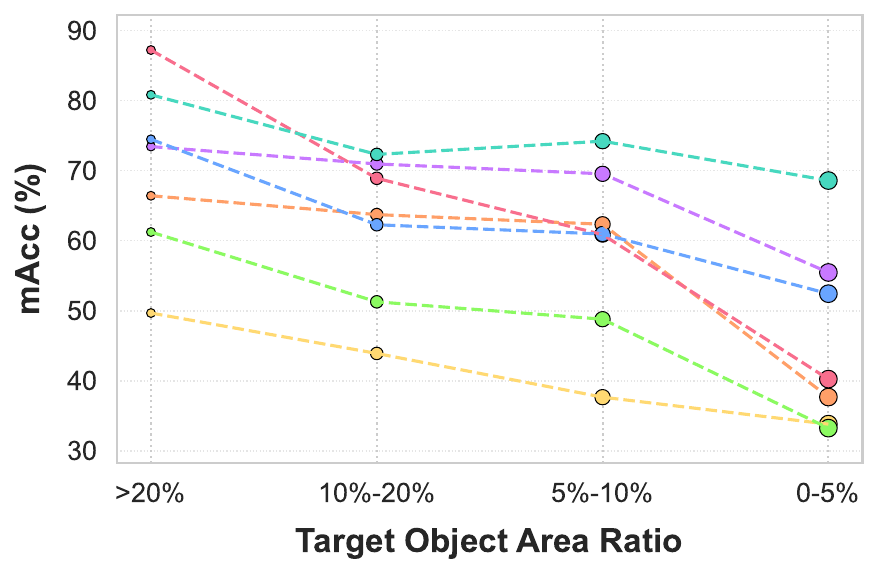}
        \caption{}
        \label{fig:bbox}
    \end{subfigure}
    \begin{subfigure}{0.33\textwidth}
        \centering
        \includegraphics[width=\linewidth]{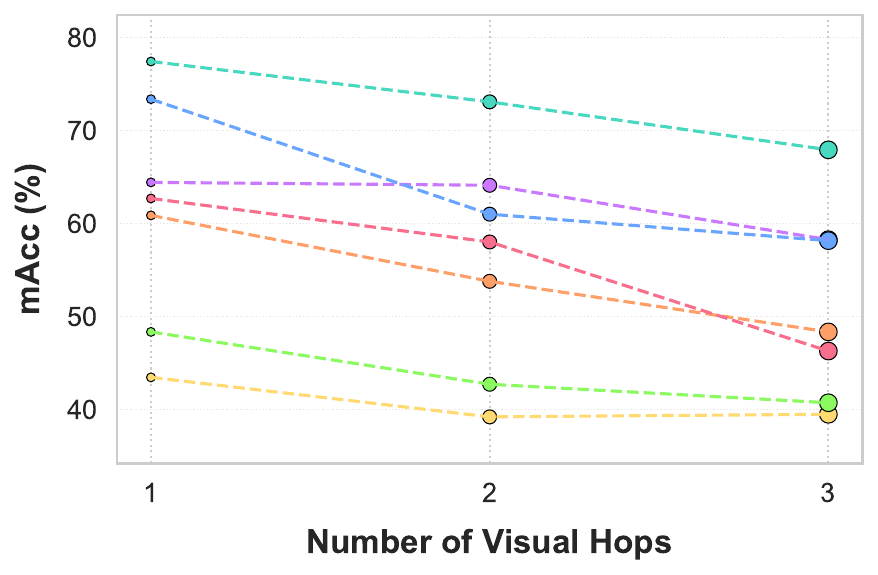}
        \caption{}
        \label{fig:hops}
    \end{subfigure}
    
    \caption{A comprehensive evaluation of model performance across three key aspects on the perception dimension: (a) the number of visual distractors that share the same category as the target, (b) the target object area ratio, and (c) the number of visual hops which is defined as the number of intermediate reference objects mentioned to locate the target .}
    \label{fig:perception_analysis}
\end{figure*}

\begin{table}[t]
\centering
\caption{ Comparisons on \bench, RefCOCO/+/g, and Ref-L4~\cite{chen2025revisiting}. RefCOCO/+/g denotes the average score across: RefCOCO~\cite{refcoco}, RefCOCO+~\cite{refcoco}, and RefCOCOg~\cite{refcocog}.}
\resizebox{1.0\columnwidth}{!}{
\begin{tabular}{l c c cc cc}
\toprule
&\multicolumn{2}{c}{\textbf{\bench}} 
& \multirow{2}{*}{RefCOCO/+/g}
& \multirow{2}{*}{Ref-L4} \\

\cmidrule(lr){2-3} 
\textbf{} & \textbf{$\text{Acc}_p$} & \textbf{$\text{Acc}_o$} 
\\
\midrule
Grounding Dino L &37.58 & 31.34 &86.59&41.75\\
VLM-R1-3B & 54.41 & 45.34 &86.23&82.56 \\
CogVLM-Grounding-17B & 57.05 & 47.54 &88.97 &81.70\\
Qwen2.5-VL-7B & 57.58 & 48.50 & 86.68 & 81.24 \\
Ovis2.5-9B & 61.74 & 51.45 &90.06&90.29\\
Qwen3-VL-8B & 71.44 & 62.16 & 89.93&81.70\\
\bottomrule
\end{tabular}
}
\label{tab:exp_compare_bmk}
\end{table}

\subsection{Benchmarking on \bench}
\noindent \textbf{Overall Performance.} 
The overall performance is summarized in Table~\ref{tab:benchmarking}.
Although most models have reached performance saturation on RefCOCO~\cite{refcoco} (Table~\ref{tab:exp_compare_bmk}), they all exhibit a substantial decline in \bench, with none exceeding  $\mathrm{Acc}_p$ of 72\%. Furthermore, a clear performance gap is observed between Visual-cue Perception ($Acc_{API}$) and reasoning in relation and commonsense ($ACC_{RC}$), with an average drop of 6.02\%. This highlights a critical limitation in current models: failure to bridge basic visual perception with compositional reasoning.
Moreover, most models lack reliable rejection capability because they are trained on positive instances and consistently generate localization hallucinations when objects are absent.



\noindent \textbf{Comparisons with Open-vocabulary Grounding Models.} 
Although open-vocabulary grounding models such as Grounding DINO have fewer parameters, they remain competitive in perceiving visual cues, achieving 0.47 on Attribute and 0.43 on Position. This performance is attributed to their training on large-scale open-vocabulary data and region–text alignment. However, when presented with referring expressions requiring compositional reasoning, their performance drops by 11\%. 
In contrast, MLLMs demonstrate superior grounding generalization across diverse referring expressions, enabled by  cross-modal understanding.

\noindent \textbf{The effectiveness of MLLMs in REC.}
Proprietary models such as GPT-5 exhibit limitations on the REC task, despite their strong general-purpose capabilities. In contrast, open-source MLLMs consistently achieve higher accuracy by incorporating visual grounding tasks to refine fine-grained perceptual understanding. Meanwhile, results for several RL-based specialist MLLMs, including VLM-R1, UniVG-R1, and Rex-Thinker, are reported.  The RL-based framework enables them to generate interpretable reasoning traces, thereby improving the faithfulness of object referring. While these methods are commonly evaluated on simpler benchmarks such as the RefCOCO series, they also achieve strong performance on \bench, particularly on tasks that require compositional reasoning. These results demonstrate that \bench provides a comprehensive benchmark fir evaluating RL-based methods in REC and holds significant potential to advance the field.


\subsection{Perceptual-aware Evaluation}
To investigate the underlying causes of unsatisfactory performance in perception of \bench, we explore three factors including: \textbf{the number of visual distractors} which is defined as objects belonging to the same category as the target, \textbf{the target object area ratio}, and \textbf{the number of object hops}, which is defined as the number of intermediate reference objects mentioned to locate the target. We divide the perception of \bench into different difficulty levels and select representative models for evaluation.  Experimental results are presented in  Figure~\ref{fig:perception_analysis}.

As shown in Figure~\ref{fig:perception_analysis}, model performance  declines  with increasing difficulty level. Under the  easiest level, most models maintain accuracy above 60\%, suggesting  that their performance on simple tasks is comparable to that  on existing benchmarks. Under the most challenging difficulty level, only Qwen3-VL-8B maintains  an accuracy above 60\%.  The average model performance declines by 20.77\%  as the number of visual distractors increases, by 25.64\% as the target  bounding box size decreases, and by 10.88\% as the number of object hops increases. These findings indicate that current models still exhibit limitations in referential grounding under visually complex scenes. 

\subsection{Reasoning-aware Evaluation}
\begin{table}[]
  \centering
  \caption{Performance of models with and without thinking mode.}
  \label{tab:exp_thinking}
  \resizebox{0.8\columnwidth}{!}{%
  \begin{tabular}{lccc}
    \toprule
    Model
    &\textbf{Relation} & \textbf{Commonsense}  \\ 
    \midrule
    Qwen3-VL-8B-Instruct &68.93&68.29\\
    Qwen3-VL-8B-Thinking &\textbf{70.41}\textcolor{red}{\textbf{($\uparrow$1.48)}}&\textbf{69.29}\textcolor{red}{\textbf{($\uparrow$1.00)}}\\
    \midrule
    Ovis2.5-9B & 58.66 & 61.04 \\
    $\text{Ovis2.5-9B}_\text{thinking}$ &\textbf{62.29}\textcolor{red}{\textbf{($\uparrow$3.63)}}&\textbf{61.71}\textcolor{red}{\textbf{($\uparrow$0.67)}} \\
    \midrule
    GLM4.1V-9B-Base &57.03& 61.83\\
    GLM4.1V-9B-Thinking &\textbf{67.89}\textcolor{red}{\textbf{($\uparrow$10.86)}}&\textbf{69.37}\textcolor{red}{\textbf{($\uparrow$7.54)}} \\
    

    \bottomrule
  \end{tabular}%
  }
\end{table}
\noindent \textbf{Effectiveness of Thinking.} We adopt three models: Qwen3-VL-8B~\cite{bai2025qwen3},  Ovis2.5-9B~\cite{lu2024ovis}, and  GLM4.1V-9B~\cite{vteam2025glm45vglm41vthinkingversatilemultimodal}, all  of which support a thinking mode, to investigate their effectiveness on Relation and Commonsense reasoning. Detailed results are  reported in Table~\ref{tab:exp_thinking}. Across all three models,  enabling the thinking mode yields consistent and substantial  improvements. 
These results indicate that explicit reasoning traces significantly improve performance on reasoning tasks, demonstrating that the construction of reliable thinking chains enhances the generalization of predictions.


\noindent \textbf{Grounding Hallucination.}
In the Rejection task, we find that the query formulation significantly affects model performance. To further evaluate whether  models possess the ability to determine the presence of a  target object, we design two evaluation settings. In the standard  setting, the model is instructed to output a bounding box if the object  exists. In the alternative setting, the task is reformulated as a binary  classification problem, where the model is asked to respond with “yes”  or “no” to indicate whether the target object is present in the image. The results, presented in Table~\ref{tab:exp_reject}, show that performance remains close to random chance (approximately 50\%) under the  binary classification setting. This indicates that models suffer from grounding hallucination  when presented with rich visual cues in image and high-density target references  in the text.  
\begin{table}[]
  \centering
  \caption{Performance comparisons across two rejection settings.}
  \label{tab:exp_reject}
  \resizebox{0.9\columnwidth}{!}{%
  \begin{tabular}{lccc}
    \toprule
    \multirow{2}{*}{Model}&
    \multirow{2}{*}{\textbf{$Acc_{RC}$}}
    &\multicolumn{2}{c}{\textbf{Rejection setting}} \\ \cmidrule(lr){3-4} 
    & &\textbf{Grounding} & \textbf{Classification} \\ 
    \midrule
  InternVL3-8B & 17.4 & 21.3 & 46.8  \\
  InternVL3.5-8B & 37.5 & - & 49.2   \\
  Qwen2-VL-7B& 43.0 & 28.5 & 51.7\\
  Qwen2.5-VL-7B& 52.3 & 3.1 & 55.1 \\
  Qwen3-VL-8B  & 68.6 & 15.8 & 64.2 \\
    \bottomrule
  \end{tabular}%
  }
\end{table}


\subsection{Ablation Study}
\begin{figure}
\includegraphics[width=\linewidth]{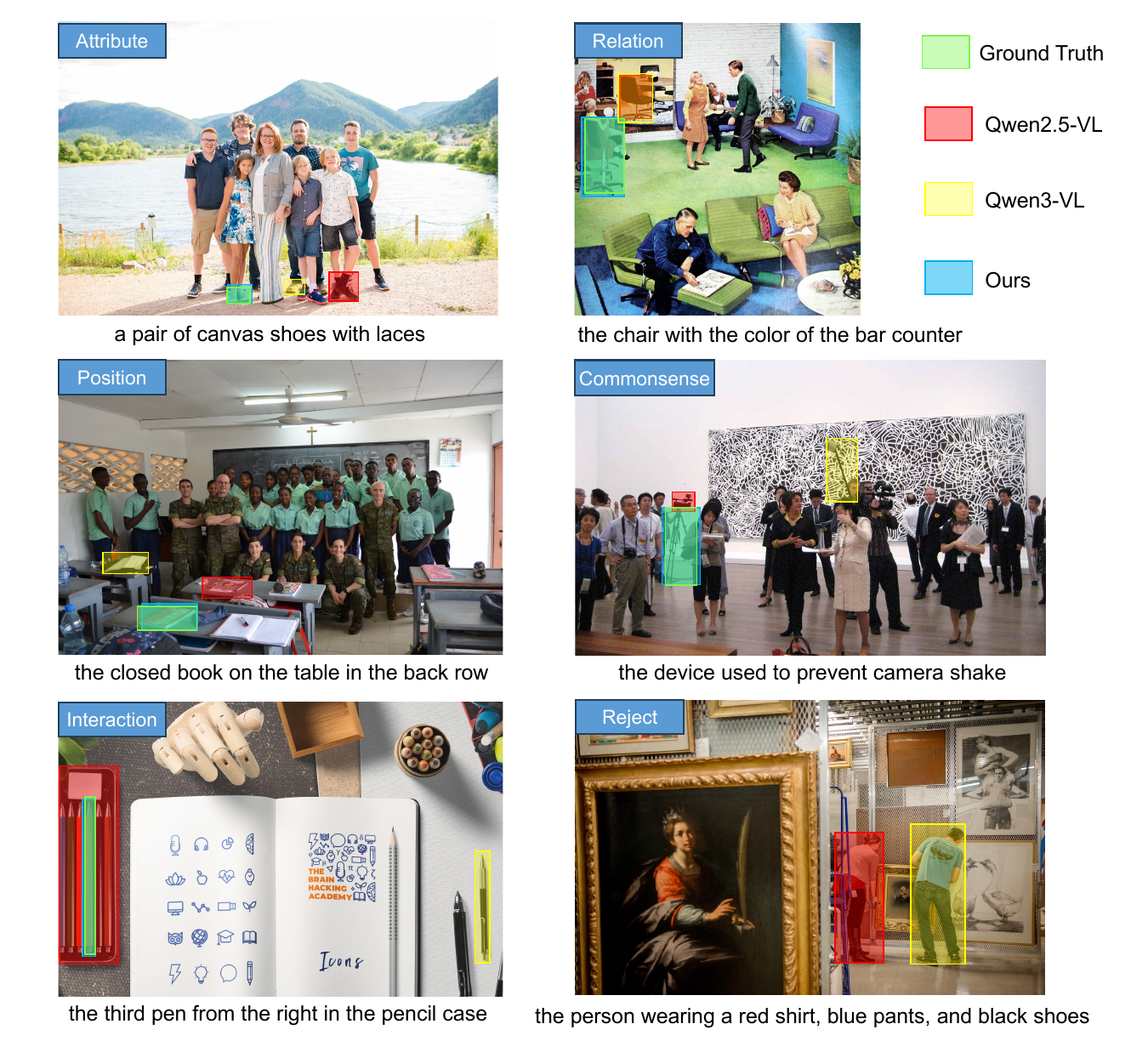}
\caption{Visual comparisons of Ref-R1 and state-of-the-art MLLMs, including Qwen2.5-VL (our baseline model) and Qwen3-VL, across the six tasks of \bench.} 
\label{fig:visualization}
\end{figure}

We investigate the impact of different components in the proposed Ref-R1 and report ablation results in Table~\ref{tab:ablation}, including evaluations on widely used benchmarks: RefCOCO/+/g~\cite{refcoco,refcocog} and Ref-L4~\cite{chen2025revisiting}. In the SFT stage, the model achieves superior performance by incorporating explicit chain-of-thought supervision. In the RL stage, we compare against VLM-R1~\cite{shen2025vlm}, which uses a fixed-threshold IoU reward. Our dynamic IoU reward outperforms this baseline. This gain arises from a dynamic threshold that progressively increases localization difficulty while placing greater emphasis on small objects, thereby encouraging attention to fine-grained visual details. Additionally, the group quality reward plays a crucial role in performance improvement. By identifying hard groups and amplifying the relative advantage of correct responses over incorrect ones within each group, this reward reinforces the model’s preference for accurate predictions. Furthermore, we present several visualizations in Figure~\ref{fig:visualization}, which demonstrate that our model, Qwen2.5-VL trained with Ref-R1, has learned to detect small objects more effectively and to integrate visual evidence with textual cues.

\begin{table}[t]
\centering
\caption{Ablation experiments of Ref-R1 on several benchmarks, where RefCOCO/+/g denotes the average score across three  datasets: RefCOCO~\cite{refcoco}, RefCOCO+~\cite{refcoco}, and RefCOCOg~\cite{refcocog}.}
\resizebox{1.0\columnwidth}{!}{
\begin{tabular}{l c c cc cc}
\toprule
\multirow{2}{*}{\textbf{Methods}}
&\multicolumn{2}{c}{\textbf{\bench}} 
& \multirow{2}{*}{RefCOCO/+/g}
& \multirow{2}{*}{Ref-L4} \\

\cmidrule(lr){2-3} 
\textbf{} & \textbf{$\text{Acc}_p$} & \textbf{$\text{Acc}_o$} 
\\
\midrule
Qwen2.5-VL & 57.58 & 48.50 & 86.68 & 81.24 \\
\midrule
SFT & 62.05 & 61.46 & 87.10 &  83.57\\
SFT-CoT & 64.11 & 63.68 & 87.02 & 84.09 \\
\midrule
GRPO$+R_{IoU}$ & 66.69 & 65.21 & 87.78 & 84.77 \\
GRPO$+R_{DyIoU}$ & 68.15 & 66.41 & 88.30 & 85.14 \\
GRPO$+R_{GroupQuality}+R_{IoU}$ & 68.07 & 66.29 & 87.96 & 85.10 \\
\midrule
\textbf{Ref-R1} & \textbf{69.35}\textcolor{red}{\textbf{($\uparrow$11.77)}} & \textbf{67.49}\textcolor{red}{\textbf{($\uparrow$18.99)}} & \textbf{88.82}\textcolor{red}{\textbf{($\uparrow$2.14)}} & \textbf{85.67}\textcolor{red}{\textbf{($\uparrow$4.43)}} \\
\bottomrule
\end{tabular}
}
\label{tab:ablation}
\end{table}




\section{Conclusion}

In this work, we observe that current REC benchmarks focus mainly on perception and do not provide interpretable score to measure how MLLMs combine perception with higher-level reasoning during referring expression comprehension. To address this, we introduce \bench, a new REC benchmark that organizes referring expressions along two core dimensions, perception and reasoning, and breaks them down into six increasingly difficult tasks: attribute, position, interaction, commonsense, relation, and rejection. 
 We also present RefObjects-200K, a large-scale referring expression dataset, generated via a fine-grained referential annotation pipeline, along with Ref-R1, a reinforcement learning–based training framework that establishes a strong baseline.
Our experiments show that \bench offers a more comprehensive and challenging evaluation for REC, and can help guide the development of MLLMs with stronger perception and reasoning abilities.

{
    \small
    \bibliographystyle{ieeenat_fullname}
    \bibliography{main}
}

\clearpage
\setcounter{page}{1}
\maketitlesupplementary

\section{Detailed Definition for Each Subcategory of \bench}
\noindent{\textbf{Attribute}}. The Attribute category focuses on the intrinsic and directly observable visual properties of objects, which includes characteristics such as an object's color, texture, material, shape, and state.

\noindent{\textbf{Position}}. The Position sub-category captures the spatial relationships between different objects within an image. The emphasis is on localizing an object based on its placement with respect to other, different kinds of objects in the scene.

\noindent{\textbf{Interaction}}. The Interaction category assesses the model's ability to understand relative relationships and arrangements among objects of the same category. Unlike the Position sub-category, this often involves ordinal references or relative positioning within a homogeneous group. 

\noindent{\textbf{Relation}}. The Relation category requires compositional reasoning, where the target object is identified by establishing a comparative link to a distinct 'anchor' object based on shared or contrasting visual attributes. Unlike simple spatial relations, this involves reasoning about properties like color, material, or texture, challenging the model to ground multiple entities and then perform a comparative judgment.

\noindent{\textbf{Commonsense}}. The Commonsense category challenges the model to identify objects based on contextual or functional descriptions rather than their explicit visual attributes. The referring expression describes an object by its purpose, potential use, or context within the scene. 

\noindent{\textbf{Reject}}. The Reject category evaluates the model's ability to handle negation and non-existent references. In these cases, the referring expression contains one or more attributes that do not match any object in the image.

\begin{table}[]
  \centering
  \caption{Detailed selection rules for target object.}
  \label{tab:selection_function}
  \Large
  \resizebox{\columnwidth}{!}{%
  \begin{tabular}{@{} l p{0.75\textwidth} @{}}
    \toprule
    \textbf{Task Category} & \textbf{Selection Rules for Target object} \\
\midrule

Attribute   & The visual context must contain multiple instances of the object's semantic category. 
The target object is selected if it possesses an intrinsic, observable property (e.g., color, texture, material) that is not shared by any other instance within that same category, thereby making the attribute a unique and fine-grained visual identifier. \\
\midrule

Position    & The target object must be uniquely localizable through an unambiguous spatial relationship (e.g., 'to the left of', 'above', 'between') with one or more distinct, \textbf{heterogeneous} anchor objects. The visual context must contain multiple instances of the object's semantic category. The presence of other same-category instances makes simple category naming insufficient, necessitating spatial reasoning. \\
\midrule

Interaction & The selection requires a group of three or more instances of the same semantic category. The target object is selected if its identity is determined by its ordinal position or relative arrangement within this \textbf{homogeneous} set (e.g., 'the third from the left', 'the one in the middle'). \\
\midrule

Relation    & This selection requires a structure for compositional, comparative reasoning. The target object is identified by establishing a relational link to a distinct 'anchor' object. This link is based on a comparison of their attributes, either through \textbf{similarity} (e.g., sharing the same color) or \textbf{contrast} (e.g., having a different material). \\
\midrule

Commonsense & An object is assigned to this category when its most salient identifier is not a direct visual property or structural relationship, but rather its function, affordance, or context-implied purpose (e.g., 'something used for cutting'). \\
\midrule

Reject      & Others. \\
    

    \bottomrule
  \end{tabular}%
  }
\end{table}

\begin{table*}[t]
\centering
\caption{Detailed results of ablation experiments of Ref-R1 on \bench.}
\resizebox{1.0\linewidth}{!}{
\begin{tabular}{l c cc cccc cccc}
\toprule
\multirow{2}{*}{\textbf{Models}}
& \multicolumn{2}{c}{\textbf{Overall}}
& \multicolumn{4}{c}{\textbf{Visual-cue Perception}} & \multicolumn{4}{c}{\textbf{Compositional Reasoning}} \\
\cmidrule(lr){2-3} \cmidrule(lr){4-7} \cmidrule(lr){8-11}
 & \textbf{$\text{Acc}_p$} & \textbf{$\text{Acc}_o$} & \textbf{Attribute} & \textbf{Position} & \textbf{Interaction} & \textbf{$\text{Acc}_{API}$} & \textbf{Relation} & \textbf{Commonsense} & \textbf{Reject} & \textbf{$\text{Acc}_{RC}$} \\
\midrule

Qwen2.5-VL & 57.58 & 48.50& 61.66 & 63.03 & 58.56  & 61.08&  49.06 &55.58 & 03.10 & 52.32  \\
\midrule
SFT & 62.05 & 61.46 & 67.38 & 67.63 & 61.50  & 65.50 & 53.46& 60.28 & 58.49 & 56.87 \\
SFT-CoT & 64.11 & 63.68& 67.43 & 68.29 & 63.95 & 66.56&  57.66 &63.22 & 61.50  & 60.44  \\
\midrule
GRPO$+R_{IoU}$ & 66.69 & 65.21& 69.13 & 69.90 & 67.27 & 68.77&  60.27 &66.89 & 57.80  & 63.58  \\
GRPO$+R_{DyIoU}$ & 68.15 & 66.41& 70.84 & 71.75 & 67.52 & 70.04& 62.96& 67.68  & 57.70  & 65.32  \\
GRPO$+R_{GroupQuality}+R_{IoU}$& 68.07 & 66.29& 71.25 & 70.40 & 68.05 & 69.90 & 63.77& 66.87  & 57.40 & 65.32  \\

Ref-R1 & 69.35 & 67.49& 72.96 & 72.50 & 68.13 & 71.20& 67.74 & 65.40 & 58.20  & 66.57  \\
\bottomrule
\end{tabular}}
\label{tab:detailed_results_ablation}
\end{table*}

\section{Details of \bench Construction}

\noindent{\textbf{Data Processing}}
We source images from the FineHARD dataset, a large-scale, high-quality dataset comprising 12 million images based on the GRiT dataset. We filtered the images based on their resolution, retaining only those within the range of 1024x1024 to 2048x2048 pixels. This criterion was established to ensure a consistent quality standard by excluding images with either insufficient visual detail or excessive computational demands.

\begin{figure*}[!b]
\centering 
\includegraphics[width=\linewidth]{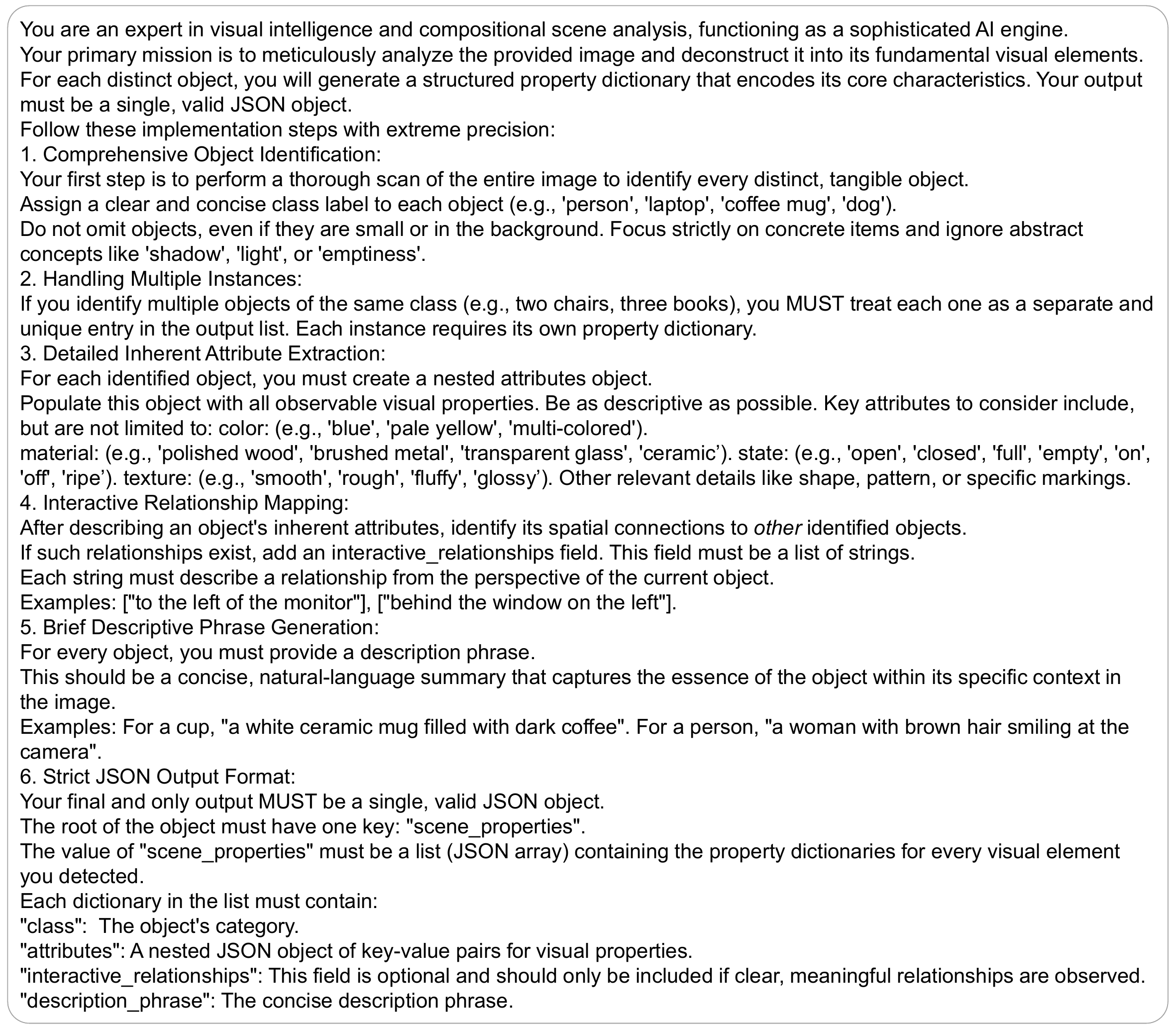}
\caption{Prompt for Image Parsing.}
\label{fig:prompt_1_description}
\end{figure*}

\begin{figure*}
\centering 
\includegraphics[width=\linewidth]{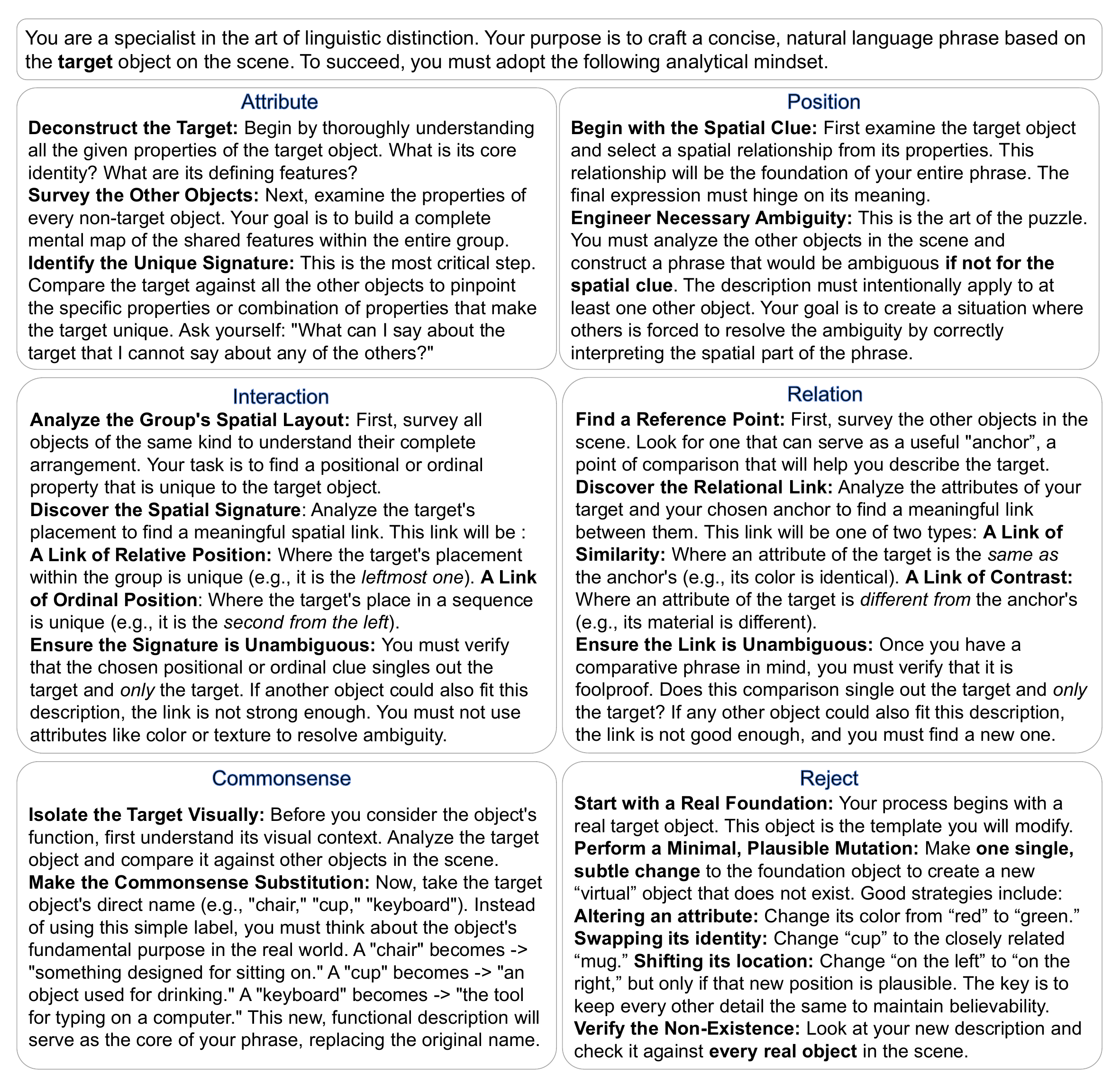}
\caption{Prompts for Referring Expression Generation.}
\label{fig:prompt_2_expression_generation}
\end{figure*}

\noindent{\textbf{Prompts Used in \bench Construction}}
Prompts used in \bench and RefObjects-200k are presented in Figure~\ref{fig:prompt_1_description} and Figure~\ref{fig:prompt_2_expression_generation}.
The prompt detailed in Figure~\ref{fig:prompt_1_description} guides the initial Image Parsing stage to deconstruct the image into structured visual elements. Subsequently, the specialized prompts in Figure~\ref{fig:prompt_2_expression_generation} are employed for the Referring Expression Generation stage to craft a unique expression for each target object based on its assigned category.

\noindent{\textbf{Task Selection Mechanism}} The detailed human-defined rule-based selection function designed to identify challenging instances for each task is shown in Table~\ref{tab:selection_function}.

\section{More Results of Ref-R1}
As shown in Table~\ref{tab:detailed_results_ablation}, Ref-R1 demonstrates comprehensive performance gains across all sub-categories. The improvements are most pronounced in Compositional Reasoning, especially within the Relation sub-category. Furthermore, the model achieves a significant uplift in the Reject task, showcasing a more robust ability to handle invalid queries.

\section{Case Study}
\begin{figure*}
\centering 
\includegraphics[width=0.8\linewidth]{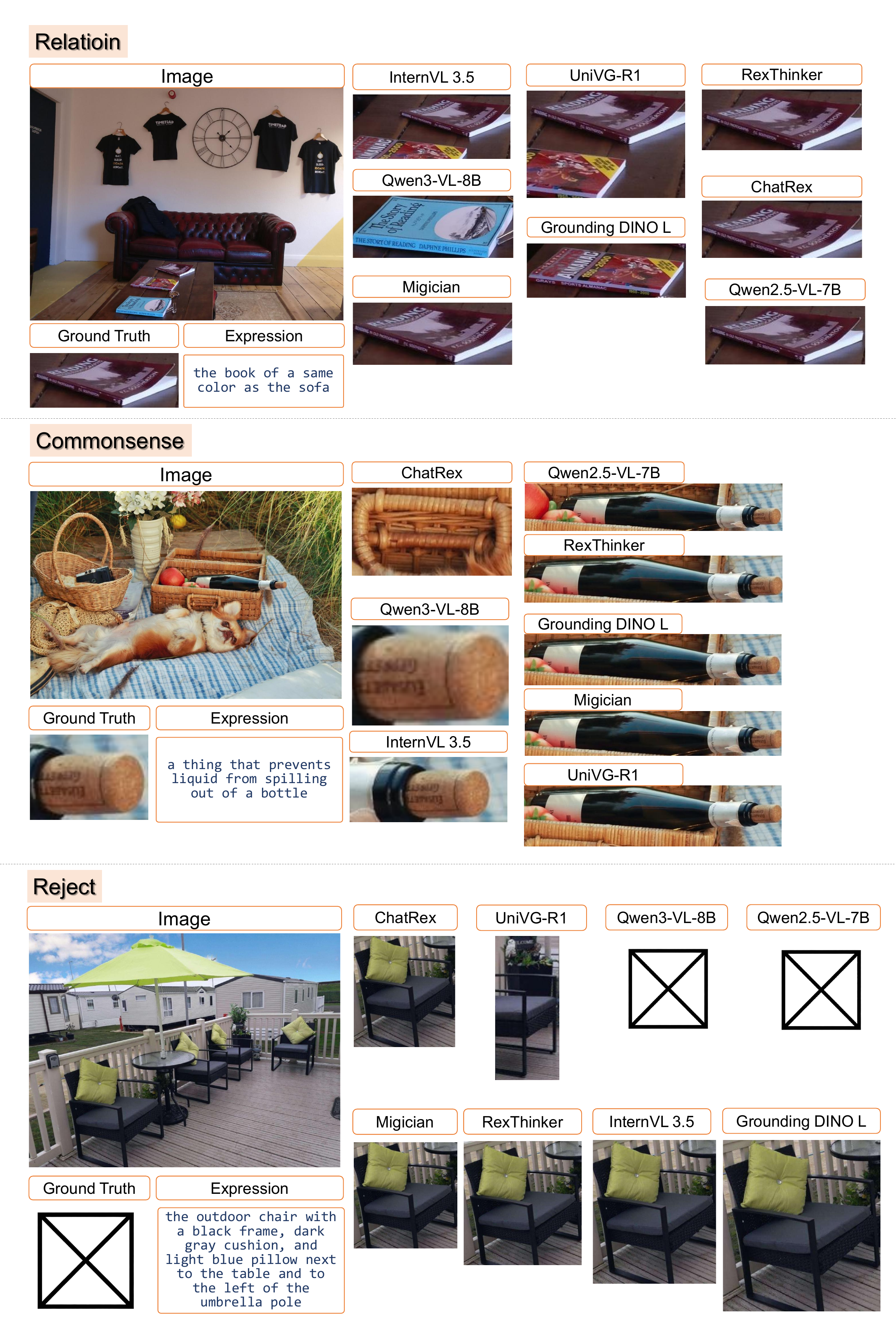}
\caption{Visual case studies for the Compositional Reasoning subcategories (Relation, Commonsense, and Reject). Each row presents a challenging example, its ground truth, and the corresponding predictions from representative models.}
\label{fig:sm_case_study2}
\end{figure*}

More cases are shown in Figure~\ref{fig:sm_case_study1} and Figure~\ref{fig:sm_case_study2}.
These figures provide a visual comparison of several representative models across challenging examples from all six categories.

\begin{figure*}
\centering 
\includegraphics[width=0.8\linewidth]{suppl/fig/sm_case_study_1.pdf}
\caption{Visual case studies for the Visual-cue Perception subcategories (Attribute, Position, and Interaction). Each row presents a challenging example, its ground truth, and the corresponding predictions from representative models.}
\label{fig:sm_case_study1}
\end{figure*}

\end{document}